\pdfoutput=1

\documentclass[11pt]{article}

\usepackage[]{acl}
\usepackage{CJKutf8} 
\usepackage{times}
\usepackage{latexsym}

\usepackage[T1]{fontenc}

\usepackage[utf8]{inputenc}

\usepackage{microtype}
\usepackage{graphicx}
\usepackage{CJKutf8}
\usepackage{amsmath}
\usepackage{bm}
\usepackage{booktabs}
\usepackage{multirow}
\usepackage{tikz}
\usepackage{pgfplots}
\usepackage{tikz-dependency}
\usepackage{graphicx}
\usepackage{enumitem}

\usepackage[center]{subfigure}
\usepackage{tikz-qtree}
\usepgflibrary{arrows.meta}
\usepgfplotslibrary{fillbetween}
\usetikzlibrary{shapes, arrows, automata, ,fit, calc, decorations.pathreplacing, positioning, shapes.geometric, backgrounds, patterns}
\usepackage{float}
\DeclareMathOperator*{\argmax}{argmax}
\setlist{nosep}

\title{MuCPAD: 
A Multi-Domain Chinese Predicate-Argument Dataset}

\author{Yahui Liu, Haoping Yang, Chen Gong\thanks{$~~$Corresponding author.}, Qingrong Xia, Zhenghua Li, Min Zhang \\ 
        Institute of Artificial Intelligence, School of Computer Science and Technology, \\ Soochow University, China \\ 
        {\tt \{yahuiliu.nlp,hpyang3,gongchen.nlp\}@foxmail.com},\\ 
        {\tt{kirosummer.nlp}@gmail.com},\\
        {\tt \{zhli13,minzhang\}@suda.edu.cn}}
 
\begin{document}
\begin{CJK*}{UTF8}{gkai}
\maketitle

\renewcommand\arraystretch{1.35}

\begin{abstract}
During the past decade, neural network models have made tremendous progress on in-domain semantic role labeling (SRL). However, performance drops dramatically under the out-of-domain setting. In order to facilitate research on cross-domain SRL, this paper presents MuCPAD, a multi-domain Chinese predicate-argument dataset, which  consists of 30,897 sentences and 92,051 predicates from six different domains. 
MuCPAD exhibits three important features. 1) Based on a frame-free annotation methodology, we avoid writing complex frames for new predicates. 2) We explicitly annotate omitted core arguments to recover more complete semantic structure, considering that omission of content words is ubiquitous in multi-domain Chinese texts. 3) We  compile 53 pages of annotation guidelines and adopt strict double annotation for improving data quality. 
This paper describes in detail the annotation methodology and annotation process of MuCPAD, and presents in-depth data analysis. We also give benchmark  results on cross-domain SRL based on MuCPAD.  

\end{abstract}


\section{Introduction}

As a fundamental NLP task,  semantic role labeling (SRL), also known as shallow semantic parsing, aims to capture the major semantic information of a sentence based on predicate-argument  structure. Basically, SRL tries to answer ``who did what to whom where and when'' \citep{marquez-2008-semantic}. Previous works have shown that SRL can help various downstream tasks, including information extraction \citep{Emanuele-jssp-ie}, plagiarism detection \citep{Paul-etal-2015-Plagiarism}, machine translation \citep{shi-etal-2016-knowledge}, reading comprehension
\citep{Zhang-etal-2020-Semantics-aware}, etc. 

Figure \ref{fig:raw-srl} gives two examples of SRL structure. 
According to the definition of semantic roles, there  exist two typical representation forms, i.e.,  the word-based and the span-based.  
This work adopts the word-based form, in which an argument corresponds to a single word. 
In contrast, span-based SRL, adopted by most previous datasets, takes a word span as an argument. The  direction of arcs is from predicates to arguments, and the labels indicate the types of semantic roles. For example, the arc from ``买 (bought)'' to ``裙子 (dress)'' with a label ``patient'' means that the semantic role between the predicate ``买  (bought)'' and the argument ``裙子 (dress)'' is ``patient''.

\definecolor{brickred}{HTML}{b92622}
\definecolor{midnightblue}{RGB}{19,41,75}
\newcommand{\white}[1]{\textcolor{violet}{#1}}
\newcommand{\blue}[1]{\textcolor{blue}{#1}}
\newcommand{\pink}[1]{\textcolor{magenta}{#1}}

\begin{figure}[tb]
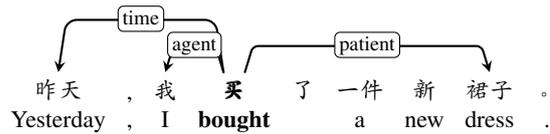
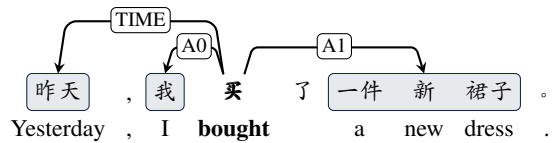

\subfigure[Word-based SRL representation adopted in MuCPAD]
{
\label{fig:zhaos-tree}
    \centering
        \begin{dependency}[]
            \begin{deptext}[row sep=-0.1cm, column sep=0.1cm, font=\small]
                         {昨天} \&{,} \& {我} \& \pmb{买} \& {了} \&{一件}\& {新} \&{裙子} \&{。}\\
                         {Yesterday} \&{,} \& {I} \& \textbf{bought} \& {} \&{a} \&{new}\& {dress} \&{.}\\
            \end{deptext}
            \depedge[edge style={thick}, edge height=4ex]{4}{1}{time}
            \depedge[edge style={thick}, edge height=2ex]{4}{3}{agent}
            \depedge[edge style={thick}, edge height=2ex]{4}{8}{patient}
        \end{dependency}
}

\subfigure[Span-based SRL representation adopted in CPB and CNB]
{
\label{fig:ours-tree}
   \centering
        \begin{dependency}[]
            \begin{deptext}[row sep=0.0cm, column sep=0.1cm, font=\small]
                         昨天 \&, \& 我 \& \pmb{买} \& 了 \&一件\& 新 \&裙子 \&。\\
                         Yesterday \&, \& I \& \textbf{bought}\&  \&a \&new\& dress \&.\\
            \end{deptext}
          \depedge[edge style={thick}, edge height=4ex]{4}{1}{TIME} 
          \depedge[edge style={thick}, edge height=2ex]{4}{3}{A0} 
          \depedge[edge style={thick}, edge height=2ex]{4}{7}{A1} 
          \wordgroup[thin, draw=midnightblue, fill=midnightblue!10]{1}{1}{1}{prd1-time}
          \wordgroup[thin, draw=midnightblue, fill=midnightblue!10]{1}{3}{3}{prd1-a0}
          \wordgroup[thin, draw=midnightblue, fill=midnightblue!10]{1}{6}{8}{prd1-a1}

        \end{dependency}
 }
\caption{Examples of two SRL formulations. 
}
\label{fig:raw-srl}
\end{figure}


Recently, Chinese SRL research has achieved tremendous  progress, thanks to the rise of deep learning methods \citep{Marcheggiani2017ASA,  He-2018-SyntaxFS, cai-2018A-srl}, especially of powerful pre-trained language models (PLMs) \citep{shi-2019-simple, conia-navigli-2020-bridging, paolini-2021-structured}. 
However, 
existing studies on Chinese SRL mainly focus on the in-domain setting, where training and test data are from the same domain \citep{ZhenWang2015ChineseSR, Guo2016AUA, Xia-2017-CSB}. SRL performance drops dramatically when the domain of test data is different from that of the training data, known as the domain adaptation problem. 

Meanwhile, with the rapid growth of user-generated web  data, cross-domain SRL has become an important and challenging task in realistic NLP systems \citep{Jiang-2007-InstanceWF, Ramponi-2020-NeuralUDA}. 
However, due to the scarcity of 
multi-domain labeled data, recent research on SRL 
makes very limited progress in the domain adaptation scenario. 

As far as we know, there are three publicly available  Chinese SRL  datasets, i.e., Chinese Proposition Bank (CPB) \citep{Xue-2007-CPB}, Chinese NomBank (CNB) \citep{Xue-2006-CNB}, and Chinese SemBank (CSB) \citep{Xia-2017-CSB}. 
All these datasets mainly contain canonical texts from newspapers or magazine/textbook articles. 


\begin{figure}[tb]
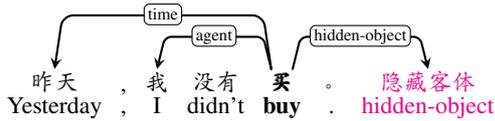

    \centering
    \begin{minipage}[t]{1\linewidth}
            \centering
            \begin{small}
            \begin{dependency}[]
                \begin{deptext}[row sep=-0.1cm, column sep=0.1cm, font=\small]
                             {昨天} \&, \& {我} \& 没有 \& {\pmb{买}}  \& 。 \& \color{magenta}{隐藏客体} \\
                             
                             {Yesterday} \&, \& {I} \& didn't  \& \textbf{buy} \& . \&  \color{magenta}{hidden-object} \\
                \end{deptext}
                \depedge[edge style={thick}, edge height=5ex]{5}{1}{time}
                \depedge[edge style={thick}, edge height=3ex]{5}{3}{agent}
                \depedge[edge style={thick}, edge height=3ex]{5}{7}{hidden-object}
            \end{dependency}
            \end{small}
        \end{minipage}%
    \caption{An example sentence with an omitted core argument in MuCPAD.}
    \label{fig:raw-srl-2}
\end{figure}



In order to facilitate research on cross-domain SRL,  this paper presents MuCPAD, a multi-domain Chinese  predicate-argument dataset, consisting of 30,897 sentences and 92,051 predicates, from 6  different domains. 
Overall, MuCPAD has the following important features.  
\begin{enumerate}[label =(\arabic*),leftmargin=*]
\item Following CSB instead of CPB and CNB, we adopt a frame-free annotation methodology,  considering that it requires a very high level of linguistic background to define new frames for new  predicates or new senses, and 
a lot of new predicates or new senses may appear in \textcolor{black}{multi-domain texts}. 
\item \textcolor{black}{As shown in Figure \ref{fig:raw-srl-2}, we explicitly annotate omitted core arguments with two special labels, i.e., ``hidden-subject'' and ``hidden-object'', in order to capture richer semantics expressed by  predicates. 
It is ubiquitous that people try to avoid repetition by omitting previous content in context, especially in non-canonical Chinese texts.
}

\item We adopt strict double annotation for all sentences in order to improve quality. If two annotators submit inconsistent results, a senior annotator determines the final answer. 
We also compile 53 pages of annotation guidelines to be studied and referred to by the annotators.  
\end{enumerate}

Based on our newly annotated MuCPAD, we conduct preliminary cross-domain SRL experiments and analysis. 
We enhance the basic SRL model by exploiting CPB2.0 as a heterogeneous dataset under the multi-task learning (MTL) framework, and by utilizing powerful  
contextualized word representations from pretrained language models (PLMs).


\textcolor{black}{We release MuCPAD along with our annotation guidelines for research usage at   \href{https://github.com/SUDA-LA/MuCPAD}{https://github.com/SUDA-LA/MuCPAD}.}
\section{Related Work}

\paragraph{English SRL Data.} 

Large-scale annotated data is a prerequisite to develop high-performance SRL systems \citep{Frstenau-2009-SemiSupervisedSRL, xia-etal-2020-semantic}. 
The most representative ones in English are FrameNet \citep{FrameNet-1998}, PropBank \citep{Kingsbury-2002-PropBank}, and NomBank \citep{Meyers-2004-NomBank}. 
FrameNet is a large-scale manually annotated semantic lexicon resource and uses semantic frames to represent meanings of words. A frame corresponds to a sense of a word, and defines the specific meanings of its core roles (i.e., ``A0-A5''). In other words, labels for core semantic roles have predicate-sense-specific  meanings. 

PropBank and NomBank are built by adding  predicate-argument structures to the constituents of syntactic parser trees in Penn Treebank \cite{Marcus-1993-treebank}. 
Their semantic roles are naturally span-based, instead of word-based. 
PropBank considers verbal predicates, while NomBank supplements nominal predicates. 
Following FrameNet, PropBank and NomBank use frames to represent semantic meanings of predicates.
However, the development of frames is both time-consuming and labor-intensive, and requires annotators to be equipped with strong linguistic background. 

The texts of PropBank and NomBank are mainly from the news domain, i.e., Wall Street Journal, except 426 sentences from the Brown corpus, which is usually used as an out-of-domain section of PropBank.


It is also noteworthy that there are PropBank-style SRL data for other languages, such as Portuguese \citep{Duran-2011-PropbankBrAB, Duran-2012-PropbankBrAB}\textcolor{black}{, Arabic \citep{pradhan-etal-2012-conll}, Finnish \citep{Haverinen-2015-TheFP}, and Turkish \citep{Sahin2018AnnotationOS}.}

\begin{table*}[htbp]
	\begin{center}
		\begin{scriptsize}
			\newcommand{\tabincell}[2]{\begin{tabular}{@{}#1@{}}#2\end{tabular}}
			\begin{tabular}{l | l | l | l }
				\hline
				& Label  & Example & Argument\\
				\hline
				\multirow{10}*{Core roles} 
				& agent & 我[打]他 (I [hit] him) & 我 (I)\\
				& co-agent  & 我和他[讨论] (I [discuss] with him) & 我 (I)、他 (him)\\
				& expe (experiencer)  & 天气真[好] (The weather [is]            really good) & 天气 (weather)\\
				& hidden-subject  & [吃]饭了吗？([Ate] ? )& 隐藏主体 (hidden-subject)\\
				& patient  & 他被[打]了 (He was [hit]) & 他 (he)\\
				& pred-patient (predicate-patient) & 他[喜欢]打篮球 (He [likes] playing basketball)& 打 (play)\\
				& dative   & [给]他书 ([Give] him a book)& 他 (him)\\
				& relative   & 这[是]他的书 (This [is] his book) & 书 (book)\\ 
				& hidden-object  & 你[吃]了吗？  (Have you [eaten]? ) & 隐藏客体 （hidden-object)\\
				& subj-obj (subject-object)   & 温度计[伸入]水中 (The thermometer is dipped into the water)& 温度计 (thermometer)\\
				\hline
				\multirow{15}*{Non-core roles} 
				& tool (instument)  & 用钢笔[写]字 ([Write] with pen)& 钢笔 (pen)\\
				& material  & 用颜料[刷]墙 ([Brush] the wall with pigment) & 颜料 (pigment)\\
				& manner  & 按计划[执行] ([Perform] according to plan) & 计划 (plan)\\
				& loc (location)  & 在学校[学习] ([Study] at school)& 学校 (school)\\
				& beg-loc (begin location)  & 从学校[出发] ([Start] from school) &学校 (school)\\
				& end-loc (end location)  & [流]入大海 ([Flow] to the ocean) & 大海 (ocean)\\
				& dir (direction)  & 向西[流] ([Flow] to the west) & 西 (west)\\
				& time  & 星期天去[打篮球] ([Play] basketball on Sunday) & 星期天 (Sunday)\\
				& beg-tm (begin time)  & 比赛七点开始[进行] (The game [starts] at seven o 'clock)  & 七点 (seven o'clock)\\
				& end-tm (end time)  &会议[开]到三点 (The meeting [runs] until three o'clock) & 三点 (three o'clock)\\ 
				& range  & 在数学上[有]天赋 ([Have] an aptitude for mathematics) & 数学 (mathematics)\\
				& cause  & 我因为爱你才[撒谎] (I [lied] because I love you) & 爱 (love)\\
				& quantity  & 我[跑]了一圈 (I [ran] a lap) &一圈 (a lap) \\
				& separated  &我们[见]过面 (We have [met]) &面 (met) \\
				\hline
			\end{tabular}
		\end{scriptsize}
	\end{center}
	\caption{Semantic role labels adopted in our guidelines. Predicates in the example sentences are marked by ``[]''. }\label{table:summary relation}
\end{table*}

\paragraph{Chinese SRL Data.} 
CPB \citep{Xue-2007-CPB}, CNB \citep{Xue-2006-CNB}, and CSB \citep{Xia-2017-CSB} are the three publicly available SRL datasets in Chinese. 
CPB and CNB, corresponding to PropBank and NomBank in English respectively, \textcolor{black}{add  predicate-argument structure of verbal predicates and  nominal predicates} into Penn Chinese Treebank \citep{Marcus-1993-treebank}. 
The semantic roles are based on pre-defined frames as well. 
Moreover, sentences in 
CPB and CNB 
mainly come from canonical texts, such Xinhua newswire, Hong Kong news, and Sinorama Magazine \citep{Hajic-2009-TheCS}.

In contrast, CSB uses general-purpose role labels, such as ``agent'' and ``patient'', and the sentences are mainly from canonical texts such as online articles and news as well. 

\paragraph{Domain adaptation.} 
Domain adaptation has been an important and challenging research topic in NLP \citep{daume-p07-feature-sharing, Ganin-15-ICML-Gan-ADV, 
Guo-2016-AUF, 
Kim-acl2017-adversary-synthetic-stale, kevin-2018-emnlp-coview, zhao-18-ICLR-Adv-multi-source}.

\citet{kim-16-fa-conling} proposed a neural shared-private model for the cross-domain slot sequence tagging task, which utilizes separate BiLSTM encoders to obtain domain-invariant and domain-specific representations, achieving significant improvements on all domains. 
\citet{acl2019-Jiachen-ner-pgn} proposed parameter generation networks for cross-domain NER. They idea is dynamically generate parameters of network modules (such as BiLSTMs) according to predicted domain distribution. 

To facilitate cross-domain Chinese dependency parsing research, \citet{Li-2019-SemisupervisedDA} proposed a large-scale multi-domain dataset for Chinese dependency parsing. They organized the NLPCC-2019 shared task on cross-domain dependency parsing \cite{Peng-2019-OverviewOT}.  \citet{Li-2019-CrossDomainTL} rank the first place in the shared task, based on a tri-training approach. 

However, possibly due to the lack of multi-domain data, research on cross-domain SRL is scarce so far. We hope our newly annotated MuCPAD can promote future research in this direction.

\section{Data Annotation}\label{section:data-annotation}
This section describe the annotation methodology and annotation process of MuCPAD in detail.



\paragraph{Annotation guidelines.} After an extensive survey of previous works on SRL data annotation, we compile 53 pages of annotation guidelines. 
We adopt 24 fine-grained general-purpose role labels to  capture the semantic relationships between predicates and arguments, as shown in Table \ref{table:summary relation}, most of which are borrowed from the guidelines of CSB.  
In particular, we introduce two special labels, i.e.,  ``hidden-subject'' and ``hidden-object'',  to explicitly annotate omitted core arguments. 
Our guidelines illustrate each label in detail using concrete examples, and are gradually 
improved according to feedback of annotators during the course of the annotation project.  

\paragraph{Data selection.}

We select the data to annotate from six domains, i.e., news, product blogs (PB), product comments (PC), web fictions (ZX), legals (LAW), and medical (MED) domains. Table \ref{data-statistic-anno} shows the data statistics.

News consists of the sentences in Chinese SemBank \citep{Xia-2017-CSB} and CoNLL-2009 Chinese dataset \citep{Hajic-2009-TheCS}.
Specifically, we choose all 10.3K sentences with 16.5K predicates from Chinese SemBank \citep{Xia-2017-CSB} and randomly select 6.7K sentences with 24.5K predicates from CoNLL-2009 Chinese dataset \citep{Hajic-2009-TheCS}. Both PB and PC are non-canonical data from Taobao\footnote{http://www.taobao.com}, where PB is from Taobao headline website, and PC is comments on products written by users.
ZX is selected from a popular Chinese fantasy novel called ``Zhuxian'' (ZX, known as ``Jade Dynasty'').
LAW is extracted from the China artificial intelligence law challenge 2018.\footnote{http://cail.cipsc.org.cn:2018/} 
MED is crawled from the medical section of People's Daily Online\footnote{http://paper.people.com.cn} and Sina.com\footnote{https://news.sina.com.cn}.

After selecting the sentences, we also need to select the concerned predicates in the sentences for annotators to annotate their corresponding arguments. 
For news domain, we directly choose the predicates in Chinese SemBank and CoNLL-2009 Chinese dataset. For other 5 domains, we choose the predicates according to several pre-defined rules which are carefully designed by considering both the dependency tree structures of the sentences and a frame dictionary extracted from the Chinese frames\footnote{https://verbs.colorado.edu/chinese/cpb/html\_frames}. For example, the root words of dependency trees are considered as predicates; the head words with ``subject'' or ``object'' dependency labels are considered as predicates; all the words that can be matched in the frame dictionary are considered as predicates.
\begin{table}[tb]
\setlength{\tabcolsep}{3pt}
\begin{center}
\begin{small}
\begin{tabular}{l l l l l l l }
\toprule
& News &PB &PC&ZX&LAW&MED \\
\midrule
\#Sent & 16,974 &  3,753 & 3,890 & 1,575 & 2,813 & 1,892  \\
\#Pred & 40,989 & 11,317 & 17,074 & 5,891 & 11,156 & 5,624 \\
\bottomrule
\end{tabular}
\end{small}
\caption{Statistics of annotated data. ``\#Pred'' and ``\#Sent'' represent the number of predicates and sentences.} 
\label{data-statistic-anno}
\end{center}
\end{table}
\paragraph{Quality Control.}
We employ 20 undergraduate students as annotators, and select 5 experienced annotators as expert annotators to handle annotation inconsistency issues.
All the annotators are paid for their work, and the salary is determined by their annotation quantity and quality. The average salary is 28 RMB per hour. 

Before real annotation, each annotator is trained for several hours to be familiar with our guidelines and our annotation tool. During the annotation process, we adopt a strict double annotation workflow to guarantee the annotation quality. Specifically, each task is randomly assigned to two annotators to annotate independently. If the submissions from the pair of two annotators are the same, the consistent answer is taken as the final answer. Otherwise, the task is assigned to a third expert annotator to decide the final answer by comparing and analyzing the inconsistent submissions.


\paragraph{Annotation tool.}
We build a browser-based annotation tool to support the double annotation workflow. For each annotation sentence, 
the annotation tool highlights the predicate in the sentence for the annotators to annotate all the arguments of the highlighted predicate. 
We also design a ``not-predicate'' checkbox in the annotation tool and ask annotators to click this checkbox to inform us when the highlighted word is out of the range of the predicate types in our guidelines.




\section{Analysis on MuCPAD}\label{sec:analysis}
In this section, we analyze the MuCPAD dataset from different perspectives 
to gain more insights.

\setlength{\tabcolsep}{5pt}
\begin{table*}[tb]
\begin{center}
\begin{small}
\begin{tabular}{l |l c c c c c c | c}
\toprule
&  &{News} &{PB}&{PC}&{ZX}&{LAW}&{MED} & {AVG}\\
\hline
\multirow{2}{*}{Consistency ratio} &{predicate-wise} & 48.86 & 57.58 & \textbf{71.23} & 48.86 & \underline{47.18} & 50.57 & 54.05\\[2pt]
 & {argument-wise}& 74.48 & 74.67 & \textbf{83.63} & 74.65  &\underline{71.49}  & 73.24  & 75.36\\[0pt]
 \hline
\multirow{3}{*}{} & \textbf{overall} & 85.86 & 82.08 & \textbf{89.40} & 83.80 & 84.57
& 85.78  & 85.25\\[2pt]
\cline{2-9}
& {agent}& \textbf{93.50}  & \underline{82.19} & 88.16  & 91.45  & 85.54  & 86.09 & 87.82 \\[2pt]
&{time}& \textbf{91.17}  & 88.98  & 85.02  & \underline{83.69}  & 85.65  & 86.87 & 86.90 \\[2pt]  
 \multirow{1}{*}{Argument-wise }&{hidden-subject}   & 88.79  & 90.61  & \textbf{94.45}  & \underline{79.65}  & 85.54  &  90.45  &  88.25 \\[2pt]   
\multirow{1}{*}{annotation accuracy} &{patient}& 87.02  & 88.36  & \textbf{90.20}  & 86.32  & \underline{85.66}  & 89.26  & 87.80\\[2pt]
\multirow{1}{*}{}&{loc}& \textbf{85.96}  & \underline{79.12} & 83.71  & 84.32  & 84.25  & 79.30 &  82.78 \\[2pt]
&{pred-patient}& \textbf{85.12}  & 84.28  & 84.08  & \underline{83.97}  &  84.11  & 84.58 &84.36 \\[2pt]     
&{expe}& \underline{81.58}  & 85.27  & \textbf{90.89}  & 86.39  & 84.20 & 87.67  & 86.00\\[0pt] 
\bottomrule
\end{tabular}
\end{small}
\caption{
Analysis on consistency ratio and accuracy. 
``AVG'' is obtained by averaging the values of the six domains. For the first major row, ``AVG'' represents the average predicate/argument-wise consistency ratios in six domains. For the second major row, ``AVG'' represents the average accuracy of overall/each label in six domains.
Boldface indicates the maximum value of each row, underline represents the minimum value of each row.
} 
\label{data-consistency-accuracy-ratio-target}
\end{center}
\end{table*}

\paragraph{Annotation consistency.}
As aforementioned, each task is assigned to two annotators. If the two submissions are inconsistent, a third expert annotator is asked to handle the inconsistency and decide the final results. The first major row in Table \ref{data-consistency-accuracy-ratio-target} shows the predicate- and argument-wise annotation consistency ratios \cite{marcus-1994-penn, 2018-Annotation-ljg} in all domains. 

The predicate-wise consistency ratio is defined as $\frac{\#\texttt{Pred}_{\texttt{annoA}\cap \texttt{annoB}}}{\#\texttt{Pred}_{\texttt{annoA}\cup\texttt{annoB}}}$, where the denominator is the total number of predicates submitted by all annotators, and the numerator is the number of predicates with consistent arguments from all annotator pairs.
We can see that the predicate-wise annotation consistency ratios in most domains are lower than 60\%. Even the highest predicate-wise consistency ratio, which is achieved in PC domain, is only 71.23\%. This means that more than a quarter of the annotation tasks need to be further checked by a third expert annotator, demonstrating the difficulty of the SRL annotation task and the importance of performing strict double annotation to guarantee data quality. 

In addition, it is worth noting that the predicate-wise consistency ratio in PC domain is much higher than that in the other five domains. 
We believe this is related to the average number of arguments per predicate. For further investigation, we calculate the average number of arguments and find the number of average arguments per predicate is the lowest in PC domain. Therefore, it is relatively easier for the annotators to recognize the arguments in PC domain. 

The argument-wise consistency ratio is defined as $\frac{\#\texttt{Arg}_{\texttt{annoA}\cap \texttt{annoB}}}{\#\texttt{Arg}_{\texttt{annoA}\cup\texttt{annoB}}}$, where the denominator is the total number of arguments submitted by all annotator pairs, and the numerator is the number of arguments that receive the same arcs and labels from the annotator pairs.
As shown in Table \ref{data-consistency-accuracy-ratio-target}, the argument-wise consistency ratios in most of the domains are lower than 75\%, except that PC achieves the highest argument-wise consistency ratio of 83.63\%.

\paragraph{Annotation accuracy.}
In the second major row of Table \ref{data-consistency-accuracy-ratio-target}, we present the argument-wise annotation accuracy. The overall argument-wise annotation accuracy is defined as $\frac{\sum_{i=1}^{n}\texttt{\#}\texttt{Arg}_{\texttt{correct}_{i}}}{2 \times \texttt{\#}\texttt{Arg}_{\texttt{gold}}}$, where the numerator is the sum of the number of correctly annotated arguments submitted by all annotators; the denominator is the total number of all gold arguments; $n$ is the number of annotators. The reason for ``$2 \times$'' in the denominator is that each task is annotated twice since it is assigned to two annotators for double annotation. The annotation accuracies in all the domains are more than 80\%, indicating that our guidelines are reasonable, which ensures the quality of annotation data. 

To gain more insights into the accuracy regarding different labels, we calculate the accuracy of 5 core labels and 2 non-core labels with high proportions for further analysis, which is shown in the third major row of Table \ref{data-consistency-accuracy-ratio-target}. The argument-wise annotation accuracy for each label is calculated by $\frac{\sum_{i=1}^{n}\texttt{\#}\texttt{Arg}_{\texttt{correct}_{i}}^{{l}}}{2 \times \texttt{\#}\texttt{Arg}_{\texttt{gold}}^{{l}}}$,
where the numerator is the sum of the number of correctly
annotated arguments with the concerned label $l$ submitted by all annotators, the denominator is the total number of all gold arguments with the concerned label $l$; $n$ is the number of annotators.
As we can see, ``hidden-subject'' achieves the highest average accuracy, demonstrating the omitted subject is easy to recognize. The lowest average accuracy is 82.78\% on ``loc'', probably because it is a non-core label with the lowest proportion of all labels and is prone to be ignored by the annotators.

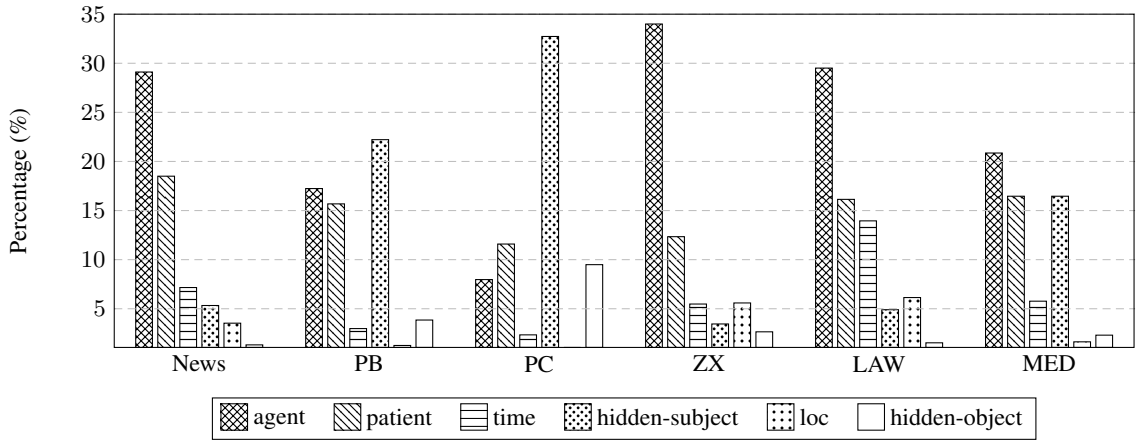
\begin{figure*}[ht]
\begin{tikzpicture}
  \centering
  \begin{small}
  \begin{axis} [    
    ybar, 
    axis on top, 
    bar width=0.22cm,
    ymajorgrids, tick align=inside,
    major y grid style={densely dashed},
    width=15cm,
    height=6cm,
    ymin=1.06, ymax=35,
    tickwidth=0pt,
    legend style={
        at={(.5,-.15)},
        legend columns=-1,
        anchor=north,
        /tikz/every even column/.append style={column sep=.3cm}
    },
    ylabel={Percentage (\%)},
    ytick={0, 5, 10, ..., 35},
        symbolic x coords={
      News, PB,  PC, ZX, LAW, MED
    },
    xtick=data,
    legend image code/.code={
        \draw [#1] (-0.15cm,-0.15cm) rectangle (0.20cm,0.20cm); 
    },
  ]
  
  \addplot [draw=black, pattern=crosshatch] coordinates {
    (News,  29.11)
    (PB,  17.24)
    (PC,  7.97)
    (ZX,   34.00)
    (LAW,   29.51)
    (MED,   20.86)
  };

  \addplot [draw=black, pattern=north west lines] coordinates {
    (News,  18.50)
    (PB, 15.68)
    (PC, 11.59)
    (ZX,   12.34)
    (LAW,  16.14)
    (MED,  16.46)
  };

  \addplot [draw=black, pattern=horizontal lines]
 coordinates {
    (News,  7.16)
    (PB,  2.98)
    (PC,  2.34)
    (ZX,   5.48)
    (LAW,   13.95)
    (MED,   5.77)
  };

  \addplot [draw=black, pattern=crosshatch dots]
  coordinates {
    (News, 5.33)
    (PB,  22.22)
    (PC,  32.73)
    (ZX,  3.45)
    (LAW,   4.9)
    (MED,   16.46)
  };

  \addplot [draw=black, pattern=dots]
coordinates {
    (News,   3.53)
    (PB,  1.26)
    (PC,  1.06)
    (ZX,  5.59)
    (LAW,   6.14)
    (MED,   1.63)
  };
 \addplot [draw=black, fill=white]
  coordinates {
    (News,  1.32)
    (PB,  3.85)
    (PC,  9.49)
    (ZX,   2.65)
    (LAW,   1.53)
    (MED,   2.31)
  };

  \legend{agent,patient,time,hidden-subject,loc,hidden-object}
  \end{axis}
  \end{small}
\end{tikzpicture}
    \caption{Label distribution in different domains. 
    }
    \label{fig:label-distribution}
\end{figure*}
\paragraph{Label distribution.}
Figure \ref{fig:label-distribution} illustrates the label distribution in the 6 domains. The labels in Figure \ref{fig:label-distribution} are sorted in descending order by their proportion in News data. We choose 2 core labels and 2 non-core labels with the highest proportions in News data. Besides, we also analyze ``hidden-subject'' and ``hidden-object'' since their distributions in different domains are specific. As shown in Figure \ref{fig:label-distribution}, the label distributions are vary across different domains. 

For PC, both ``hidden-subject'' and ``hidden-object'' account for the largest proportion compared with that for all the other 5 domains, which means PC contains the most omitted core arguments. The reason is that PC is user-generated comments on a concerned product, and people tend to directly write the comment of the concerned product and omit the product name and the personal pronoun. For example, in the sentence ``很~(very)~喜欢~(like), 买~(buy)~了~好~(very)~多~(much)'', both the hidden-subjects and the hidden-objects of the predicates ``喜欢~(like)'' and ``买~(buy)'' are omitted.
The omitted core arguments in PB also take relatively large proportion, similar to the reason for PC.


For ZX, it has the most ``agent'' roles and the least ``hidden-subject'' roles compared with other domains. This is owing to the genre of ZX texts, which are extracted from a popular Chinese fantasy novel with a lot of fictional characters. In order to make the story more understandable by readers, the names of the fictional characters are often explicitly written in the sentences, leading to more ``agent'' roles and fewer ``hidden-subject'' roles.



For LAW, it has more ``time'' roles and ``loc'' roles than other domains, 
since the elements (i.e., time and location) of the cases are usually frequently occurred to provide more accurate information.


For MED, the number of ``hidden-subject'' label accounts for a large proportion among all the labels in MED, only fewer than that of ``agent'' label,
mainly because the descriptions of symptoms in MED usually omit the subjects. For example, in the sentence ``酒精 (alcohol) 中毒 (poisoning): 发生 (occur) 昏迷 (coma) 不能 (cannot) 催吐 (induce vomiting), the subjects of the predicates ``中毒 (poisoning)'', ``发生 (poisoning)'', ``昏迷 (coma)'', ``不能 (cannot)'' are all omitted.

Looking into the distribution of ``hidden-subject'' and ``hidden-object'' labels in all the domains, we find that hidden labels exist in all the domains, especially in non-canonical texts like PC and PB, demonstrating the necessity of annotating hidden labels. 
In addition, ``hidden-subject'' takes a higher proportion than ``hidden-object'' in all the 6 domains, reflecting that the subject of the predicate in Chinese sentences is often omitted.

\paragraph{Annotation difficulties.} 
To understand difficulties during annotation, we calculate the proportion of the arguments with the same arcs but different labels from two annotators among all the arguments with the same arcs.
We find that the confusion pattern ``{agent, expe}'' accounts for the largest proportion of 22.23\%, which means the label ``agent'' is prone to be confused with ``expe''.
This is possibly because the POS for some predicates is subtle and vague in Chinese, causing the confusion of the argument labels.
Taking the sentence ``纽扣 (Buttons) 一天 (a day)  坏 (getting broken) 一个 (one)'' as an example, ``坏 (getting broken)'' may be misunderstood as an adjective and thus the argument ``纽扣 (buttons)'' is incorrectly annotated as ``expe''. Actually, ``坏 (getting broken)'' acts as a verb in this sentence and the correct label of ``纽扣 (buttons)'' is ``agent''.
The second confusion pattern is ``{patient, pred-patient}'', with a proportion of 12.6\%, due to the misunderstanding of the POS of the argument.
It is also difficult for annotators to distinguish ``agent'' and ``patient''. For example, in the sentence ``新 (new) 衣服 (cloths) \underline{被 (was)} 弄脏了 (soiled)'', the preposition ``被 (was)'' is omitted. As a result, the label of ``新 (new) 衣服 (cloths)'', which is ``patient'', may be confused with ``agent'' due to the omission.

\section{Approach}
Based on our newly annotated multi-domain Chinese SRL data, we conduct preliminary experiments, aiming to provide benchmark results.
Specifically, we present a simple basic SRL model and enhance the model with the contextualized word representations from BERT for further improvements.
Besides, we also present a MTL framework to improve the SRL performance by learning from multiple heterogeneous datasets simultaneously \citep{conia-etal-2021-unifying}. 

In this work, we focus on the \emph{predicate-given} setting, which means we do argument identification and classification according to the given predicates in one sentence.

Following previous works \citep{cai-2018A-srl, zhang-etal-2019-syntax}, we treat the predicate-given SRL task as a word pair  classification problem and try to find the predicate-argument structure $\hat{\bm{y}}$ with the highest score: 
\begin{equation}
\hat{\bm{y}} = \argmax_{\bm{y} \in \mathcal{Y}(\bm{x})} score(\bm{x}, \bm{y})
\end{equation}
where $\mathcal{Y}(\bm{x})=\{(i,j,r)| i\in \mathcal{P}, 1\leq j \leq n, r \in \mathcal{R}\}$ represents the set of all possible predicate-argument pairs.
$\mathcal{P}$ is the set of given predicates, $n$ is the number of sentence, and 
$\mathcal{L}$ is the semantic role label set, which contains 24 semantic role labels and an extra ``None'' label to indicate there is no semantic relationship between  the given predicate and the $j$-th word. 

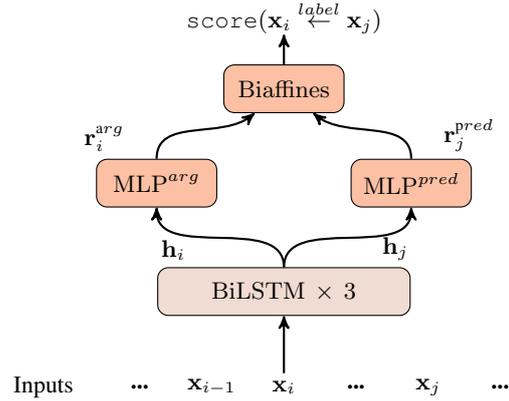
\begin{figure}[tb]
\begin{center}
\begin{small}
\begin{tikzpicture}[node distance = 0.1cm, auto]
\definecolor{mycolor}{RGB}{228, 198, 208}
\definecolor{yhcolorgray}{gray}{0.95}
\definecolor{yhcolorgreen}{rgb}{1,0.5,0}
\definecolor{mycolor2}{RGB}{161, 175, 201}
\definecolor{mycolor3}{RGB}{135, 186, 191}
\node [inner sep=0pt] (the_orig) {};
\node [inner sep=0pt, above of=the_orig, node distance = 0.5em] (xi) {};
\node [inner sep=0pt, above of= the_orig, node distance = 0.5em] (x2) {\bf $\mathbf{x}_{i}$};

\node [inner sep=0pt, left of=x2, node distance = 3em] (x1) {\bf $\mathbf{x}_{i-1}$};

\node [inner sep=0pt, left of=x1, node distance = 3em] (x0) {\bf ...};

\node [inner sep=0pt, left of = x0, node distance = 4em] (input_text) { Inputs };

\node [inner sep=0pt, right of=x2, node distance = 3em] (x3) {\bf ...};

\node [inner sep=0pt, right of=x3, node distance = 3em] (x4) {\bf $\mathbf{x}_{j}$};

\node [inner sep=0pt, right of=x4, node distance = 3em] (x5) {\bf ...};

\node [rectangle, above of =x2, node distance = 4em, rounded corners, draw, fill={rgb,255:red,230; green,197; blue,180}, fill opacity=0.6, text opacity=1.0, text centered, minimum width=10.5em, minimum height=2em] (lstm) {\bf $\mathrm{BiLSTM}\, \times \, 3$};
\node [inner sep=0pt, above left = 0.3em and -1.2em of lstm] (hi) {$\mathbf{h}_{i}$};
\node [inner sep=0pt, above right = 0.3em and -1.2em of lstm] (hj) {$\mathbf{h}_{j}$};
\node [inner sep=0pt, above of = x2, node distance = 0.5em] (above_xi){};
\path [draw, thick, ->, >=stealth'] (above_xi) to [out=90,in=-90]  (lstm);
\node [rectangle, above left = 2.5em and -2.5em of lstm, rounded corners, draw, fill={rgb,255:red,251; green,153; blue,104}, fill opacity=0.6, text opacity=1.0, text centered, minimum width=5em, minimum height=2em] (mlp_d) {\bf $\mathrm{MLP}^{arg}$};
\node [rectangle, above right = 2.5em and -2.5em of lstm, rounded corners, draw, fill={rgb,255:red,251; green,153; blue,104}, fill opacity=0.6, text opacity=1.0, text centered, minimum width=5em, minimum height=2em](mlp_h) {\bf $\mathrm{MLP}^{pred}$};

\path [draw, thick, ->, >=stealth'] (lstm) to [out=90,in=-90]  (mlp_d);
\path [draw, thick, ->, >=stealth'] (lstm) to [out=90,in=-90]  (mlp_h);

\node [inner sep=0pt, above left = 0.3em and -1.2em of mlp_d] (ri) {$\mathbf{r}_{i}^{\textup arg}$};
\node [inner sep=0pt, above right = 0.3em and -1.2em of mlp_h] (rj) {$\mathbf{r}_{j}^{\textup pred}$};

\node [rectangle, rounded corners, draw, fill={rgb,255:red,251; green,153; blue,104}, fill opacity=0.6, text opacity=1.0, text centered, minimum width=5em, minimum height=2em, above of =  lstm, node distance = 8.5em] (biaffine) {\bf $\mathrm{Biaffines}$};
\node [inner sep=0pt, above of = biaffine, node distance = 3.0em] (score) {$\texttt{score}(\mathbf{x}_{i} \stackrel{label}\leftarrow \mathbf{x}_{j})$};
\path [draw, thick, ->, >=stealth'] (biaffine) to [out=90,in=-90] (score) ;
\path [draw, thick, ->, >=stealth'] (mlp_d) to [out=90,in=-140]  (biaffine);
\path [draw, thick, ->, >=stealth'] (mlp_h) to [out=90,in=-40]  (biaffine);
\end{tikzpicture}
\caption{The architecture of our basic SRL model.}\label{fig:base-model}
\end{small}
\end{center}
\end{figure}
        
\subsection{Basic SRL Model}
Inspired by previous works \cite{cai-2018A-srl, zhang-etal-2019-syntax}, we build a basic SRL model that utilizes the biaffine attention mechanism \cite{dozat2016deep} to score each candidate predicate-argument pair.
Figure \ref{fig:base-model} shows the architecture of the basic model. 
During both training and evaluation, multiple predicates in the same sentence are handled simultaneously. First, the input sentence is encoded; then, scores between predicates and all other words are computed; finally, the roles of each predicate are determined via local classification.  

\textbf{The input vector} is the concatenation of the pre-trained word embedding $\mathbf{e}_{i}^{pre}$, the randomly initialized word embedding $\mathbf{e}_{i}^{r}$, the character-based  word representation $\mathbf{r}_{i}^{c}$, and the  predicate indicator embedding $\mathbf{e}_{i}^{p}$. 
$$\mathbf{x}_i =  \mathbf{e}_{i}^{pre}\oplus \mathbf{e}_{i}^{r} \oplus  \mathbf{r}_{i}^{c} \oplus  \mathbf{e}_{i}^{p}$$
where $\mathbf{r}_{i}^{c}$ is produced by CNN, and the Boolean predicate indicator is true only for words that are given predicates. 

\textbf{A three-layer BiLSTM} is applied to obtain  context-aware representation of each word, i.e., $\mathbf{h}_{i}$. 

\textbf{Two separate MLPs} are applied over $\mathbf{h}_{i}$ to get two lower-dimensional representation $\mathbf{h}_{i}^{pred}$ (as predicate) and $\mathbf{h}_{i}^{arg}$ (as candidate argument).

\textbf{Biaffines} are used to compute scores of  labels between a predicate and a word. 

During training, we adopt the local cross-entropy loss. 
To obtain cross-domain results on the basic SRL model, we train the model on source domain data and make predictions on target domain data.

\subsection{Enhancing with BERT}
Recently proposed PLMs, 
such as BERT \citep{Devlin-2019-BERT}, 
have shown the great power in learning and capturing  contextualized representations and have proven to be beneficial in a  variety of NLP tasks, such as information retrieval \citep{Yang-2019-SimpleAO}, question answering \citep{Yang-2019-EndtoEndOQ}, and word segmentation \citep{Hung-2020-TowardsFA}.
In this work, we extract the fixed contextualized representations from BERT for words and treat them as additional features to augment the input representation, i.e., $\mathbf{x}_i =  \mathbf{e}_{i}^{pre}\oplus \mathbf{e}_{i}^{r} \oplus    \mathbf{r}_{i}^{c} \oplus  \mathbf{e}_{i}^{p} \oplus  \mathbf{e}_{i}^{BERT} $.

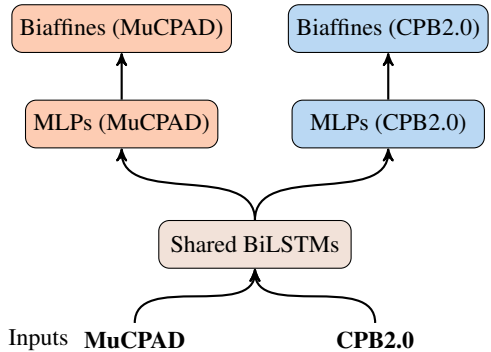
\begin{figure}[tb]
\begin{center}
\begin{small}
\begin{tikzpicture}[node distance = 0.1cm, auto]
\definecolor{mycolor}{gray}{0.95}
\definecolor{mycolor2}{rgb}{1,0.5,0}

\node [inner sep=0pt] (the_orig) {};


\node [inner sep=0pt, above of=the_orig, node distance = 0.5em] (xi) {};
\node [inner sep=3pt, left of=xi, node distance = 5em] (x0) {\bf MuCPAD};
\node [inner sep=3pt, right of=xi, node distance = 5em] (x1) {\bf CPB2.0};
\node [inner sep=0pt, left of = x0, node distance = 4em] (input_text) { Inputs };



\node [rectangle, above of =xi, node distance = 4em, rounded corners, draw, fill={rgb,255:red,230; green,197; blue,180}, fill opacity=0.5, text opacity=1.0, text centered, minimum width=8em, minimum height=2em] (lstm) {Shared BiLSTMs};

\node [rectangle, above left = 3em and -2.5em of lstm, rounded corners, draw, fill={rgb,255:red,251; green,153; blue,104}, fill opacity=0.5, text opacity=1.0, text centered, minimum width=8em, minimum height=2em] (mlp_src) {MLPs (MuCPAD)};

\node [rectangle, above right = 3em and -2.5em of lstm, rounded corners, draw, fill={rgb,255:red,117; green,178; blue,231}, fill opacity=0.5, text opacity=1.0, text centered, minimum width=8em, minimum height=2em] (mlp_tgt) {MLPs (CPB2.0)};

\node [rectangle, rounded corners, draw, fill={rgb,255:red,251; green,153; blue,104}, fill opacity=0.5, text opacity=1.0, text centered, minimum width=3.5em, minimum height=2em, above of = mlp_src, node distance = 4em] (biaffine_src) {Biaffines (MuCPAD)};

\node [rectangle, rounded corners, draw, fill={rgb,255:red,117; green,178; blue,231}, fill opacity=0.5, text opacity=1.0, text centered, minimum width=3.5em, minimum height=2em, above of = mlp_tgt, node distance = 4em] (biaffine_tgt) {Biaffines (CPB2.0)};

\node [inner sep=0pt, above of = xi, node distance = 0.5em] (above_xi){};
                
\path [draw, thick, ->, >=stealth'] (x0) to [out=90,in=-90]  (lstm);
\path [draw, thick, ->, >=stealth'] (x1) to [out=90,in=-90]  (lstm);

\path [draw, thick, ->, >=stealth'] (lstm) to [out=90,in=-90]  (mlp_src);
\path [draw, thick, ->, >=stealth'] (lstm) to [out=90,in=-90]  (mlp_tgt);

\path [draw, thick, ->, >=stealth'] (mlp_src) to [out=90,in=-90]  (biaffine_src);
\path [draw, thick, ->, >=stealth'] (mlp_tgt) to [out=90,in=-90]  (biaffine_tgt);

\end{tikzpicture}
\caption{The framework of MTL.}\label{fig:mtl}
\end{small}
\end{center}
\end{figure}

\begin{table*}[tb]
\setlength{\tabcolsep}{2pt}
\begin{center}
\begin{small}
\begin{tabular}{l l l l l l l | l}
\toprule
(\#Pred / \#Sent)& \multicolumn{1}{c}{Source } &\multicolumn{1}{c}{ PB }&\multicolumn{1}{c}{PC}&\multicolumn{1}{c}{ZX}&\multicolumn{1}{c}{LAW}&\multicolumn{1}{c}{MED}&\multicolumn{1}{|c}{CPB2.0} \\
\midrule

Train & 32,790 / 13,022 & \multicolumn{1}{c}{--} & \multicolumn{1}{c}{--} & \multicolumn{1}{c}{--} & \multicolumn{1}{c}{--} & \multicolumn{1}{c|}{--} & 72,616 / 13,170\\[2pt]
Dev & 4,098 / 1,875 & 3,796 / 1,255 & 5,658 / 1,295  & 1,784 / 492 & 3,718 / 778 & 1,874 / 478 & \multicolumn{1}{c}{--} \\[2pt]
Test & 4,101 / 2,077& 7,521 / 2,498& 11,416 / 2,595& 4,107 / 1,083 & 7,438 / 2,035 & 3,750 / 1,414 & \multicolumn{1}{c}{--}\\[0pt]


\bottomrule
\end{tabular}
\end{small}
\caption{Statistics of MuCPAD and CPB2.0. ``\#Pred'' and ``\#Sent'' represent the number of predicates and sentences.} 
\label{data-statistic-target}
\end{center}
\end{table*}


\subsection{
Utilizing Heterogeneous Data
with MTL} 
MTL is a commonly used method to improve the model performance by learning the underlying knowledge from multiple related tasks or datasets \citep{Collobert-2008-AUA, Guo-2016-AUF, Li-2019-SemisupervisedDA}. 
In this work, we design a MTL framework to utilize heterogeneous SRL datasets to boost the  SRL model performance.

\begin{table*}[tb]
\setlength{\tabcolsep}{4pt}
\begin{center}
\begin{small}
\begin{tabular}{l  *{2}{c} *{2}{c} *{2}{c} *{2}{c} *{2}{c} *{2}{c} | c} 
\toprule
\multirow{2}{*}{} 
& \multicolumn{2}{c}{ Source }
& \multicolumn{2}{c}{ PB }
& \multicolumn{2}{c}{ PC }
& \multicolumn{2}{c}{ ZX }
& \multicolumn{2}{c}{ LAW }
& \multicolumn{2}{c|}{ MED }
& \multirow{2}{*}{AVG}\\
& dev & test & dev & test & dev & test & dev & test & dev & test & dev & test     \\
\hline
Baseline &  69.55 & 68.40 & 44.24 & 44.59 & 46.12 & 47.37 & 50.98 &	49.26 & 47.05	& 49.77 &	44.88 &	46.70 & 50.74\\

Baseline+BERT & 80.25 &	79.46 & 64.28 &	\textbf{66.23} &	66.77 &	67.14 &	71.65 &	\textbf{71.77} & 70.47 &	73.28 &	66.06 &	67.97 & 70.44\\
Baseline+MTL & 74.55	& 73.43 & 46.95 & 47.36	& 48.61 & 48.95 & 54.38 & 53.54 & 48.55 & 51.25 &	54.53 &	55.26 & 54.78\\
Baseline+MTL+BERT & \textbf{81.05}	& \textbf{80.85} & \textbf{64.35}	& 65.27	& \textbf{67.75}	& \textbf{68.13}	& \textbf{72.44} &	71.72 & \textbf{70.53} &	\textbf{73.79} &	\textbf{66.38} &	\textbf{69.07 }& \textbf{70.94}\\
\bottomrule
\end{tabular}
\end{small}
\caption{F1 scores of different models on MuCPAD. ``AVG'' is obtained by averaging the values of both dev and test in all domains.} 
\label{tbl:baseline-mtl}
\end{center}
\end{table*}

As shown in Figure \ref{fig:mtl}, we extend the basic SRL model to the MTL framework.
Specifically, the SRL parsing on MuCPAD data and CPB2.0 data are considered as two separate tasks. They share the same word/predicate embeddings and BiLSTM parameters. Over the shared BiLSTMs, two separate MLPs and biaffines are employed for MuCPAD and CPB2.0 SRL parsing respectively.

\section{Experiments}

\paragraph{Data.} 
\textcolor{black}{Our experiments mainly focus on zero-shot single-source domain adaptation, that is, we have labeled training data for the source domain, and do not have labeled training data for the target domain. 
Specifically, we use the News domain of MuCPAD as the source domain, and the other five domains as target domains.}
The data statistics for source and target domains are shown in Table \ref{data-statistic-target}.
For the auxiliary data used in the MTL framework, we randomly select 13,170 sentences with 72,616 predicates from CPB2.0 \cite{Xue-2006-SemanticRL}, which belongs to the same newswire genre with the source domain data.


\paragraph{Evaluation metric.} 
We adopt the standard precision ($\frac{\#\texttt{Arg}_{\texttt{correct}}}{\#\texttt{Arg}_\texttt{pred}}$), recall ($\frac{\#\texttt{Arg}_{\texttt{correct}}}{\#\texttt{Arg}_\texttt{gold}}$), and F1 score ($\frac{2PR}{P+R}$) for SRL evaluation.



\paragraph{Settings.}
We implement the basic SRL model and MTL framework with PyTorch\footnote{https://pytorch.org/} and mainly follow the hyperparameters of \citet{cai-2018A-srl}, such as the dimensions of embeddings, learning rate, and dropout ratios. We use bert-base-chinese\footnote{https://huggingface.co/bert-base-chinese} to obtain contextualized representations for words, and the dimension of the BERT representations is 768.
During training, early stopping is triggered if the peak performance in dev data does not increase in 50 consecutive iterations.


\textbf{Results of the basic model.}
The first row of Table \ref{tbl:baseline-mtl} presents the results in the source/target domain dev/test data using the basic SRL model trained on the source data.

First, it is obvious that the performance in all the five target domains drops dramatically compared with the results on source data, with the gap of more than 18\% in F1. This indicates that the model trained on source data has a challenge in making reliable predictions on target domain data due to the distributional mismatch between different domains. 
Second, we find that the basic SRL model performs better on ZX and LAW compared with the other three target domains data, i.e., PB, PC, and MED. The probable reason is that ZX and LAW are novel and legal case, respectively, which are more canonical in text. 
Third, PB has the lowest F1 score in both dev and test. This can be explained by the fact that PB is non-canonical data from Taobao headline website. The dissimilarity between the source training data and PB target data causes the low performance.

\textbf{Results with BERT.}
The second row of Table \ref{tbl:baseline-mtl} shows the results of the baseline with BERT representations.
We can see that the results of ``Baseline+BERT'' consistently increase by large margins compared with the corresponding baseline models without BERT (as shown in the first major row of table \ref{tbl:baseline-mtl}), demonstrating the great power of BERT in contextualized representation.

\textbf{Results with heterogeneous CPB2.0.}
As shown in the third row of Table \ref{tbl:baseline-mtl}, benefiting from the additional semantic information provided by the auxiliary CPB2.0 data using the MTL framework, the SRL performance in all domains are improved compared with the baseline model. This indicates that the MTL framework is effective in capturing and learning the underlying common knowledge from heterogeneous data.

On the one hand, comparing the improvements brought by MTL in all domains, we find that MED data obtains the largest gains of 9.65\%/8.56\% F1 in dev/test, respectively.
The main reason is that the MED data belongs to the same newswire domain as the auxiliary CPB2.0 data.
On the other hand, the improvement in LAW is the smallest. 
This can be explained by the difference in label distribution between \textcolor{black}{LAW and CPB2.0. For example, as mentioned in Section \ref{sec:analysis}, the labels ``time'' and ``loc'' in LAW account for the largest proportion (13.95\% and 6.14\% respectively) compared with other domains. However, the proportions of ``time'' and ``loc'' in CPB2.0 data are only 6.10\% and 3.40\% respectively} (about half of that in LAW). Therefore, CPB2.0 cannot provide much more valid information to increase the performance of these labels.

\textbf{Results with BERT and heterogeneous CPB2.0.}
Finally, when utilizing both BERT representations and the heterogeneous CPB2.0 data on our baseline, the enhanced model gives the best or comparable results in 5 of the 6 domains, with an average increase of 0.5\% F1, showing that the MTL framework is effective in utilizing heterogeneous data and can complement the information obtained from BERT representations.





\section{Conclusions}
This paper presents a multi-domain Chinese predicate-argument dataset, named MuCPAD, which consists of 30,897 sentences with 92,051 predicates and covers 6 different domains. 
In particular, we adopt a frame-free annotation  methodology, which does not require high-level linguistic background for defining frames for large amounts of new predicates or new senses in multi-domain data. 
Besides, considering that omission of content words is  ubiquitous in Chinese, we explicitly annotate omitted core arguments with two special designed labels ``hidden-subject'' and ``hidden-object'' 
for better semantic understanding. 
To ensure annotation quality, we adopt strict double  annotation and ask a third expert to handle annotation inconsistency.
We also perform analysis on MuCPAD from different perspectives. 
Finally, we conduct preliminary cross-domain experiments and analysis on MuCPAD. 


\section*{Acknowledgements}

The authors would like to thank the anonymous reviewers for the helpful comments. We are greatly grateful to all our annotators for their hard work in data annotation. This work was supported by the National Natural Science Foundation of China (Grant No. 62176173, 61876116), and a project funded by the Priority Academic Program Development of Jiangsu Higher Education Institutions.

\bibliography{anthology,custom}


\end{CJK*}
\end{document}